\documentclass{article}
\usepackage[nonatbib, preprint]{neurips_2026}

\usepackage[utf8]{inputenc}
\usepackage[T1]{fontenc}
\usepackage{hyperref}
\usepackage{url}
\usepackage{booktabs}
\usepackage{amsfonts}
\usepackage{amsmath}
\usepackage{amsthm}
\usepackage{nicefrac}
\usepackage{microtype}
\usepackage{xcolor}
\usepackage{graphicx}
\usepackage{algorithm}
\usepackage{algorithmic}
\usepackage[small]{caption}
\usepackage{subcaption}
\usepackage{multirow}
\usepackage{stfloats}
\usepackage{latexsym}
\usepackage[square,numbers]{natbib}
\usepackage{array}
\title{Periodic Asynchrony: An On-Policy Approach for Accelerating LLM
  Reinforcement Learning}

\author{%
  Jian Lu \\
  Big Data \& AI Lab \\
  Industrial and Commercial Bank of China (ICBC) \\
  \texttt{janelu@live.cn}
   \And
   Yi Luo\thanks{Corresponding author.} \\
   Software Development Center \\
   Industrial and Commercial Bank of China Limited \\
   \texttt{luoy@sdc.icbc.com.cn}
}

\begin{document}

\maketitle


\begin{abstract}
Since the introduction of the GRPO algorithm, reinforcement learning~(RL)
has attracted increasing attention for LLM post-training, yet training
efficiency remains a critical challenge.
In mainstream RL frameworks, inference and training are co-located on the
same devices, and their synchronous execution prevents concurrent inference
and training.
In this work, we revisit the strategy of separating inference and training
deployment, and propose a \emph{periodically asynchronous} framework that
transforms synchronous RL training into an asynchronous
producer--consumer pipeline.
By synchronising model weights at the beginning of each training iteration
and generating all rollouts from the same policy, the proposed framework
remains inherently \emph{on-policy}---without any modification to standard
RL algorithms---thereby avoiding the off-policy bias introduced by existing
asynchronous approaches.
We further introduce a unified tri-model architecture and a shared-prompt
attention mechanism to support efficient asynchronous execution and reduce
redundant computation.
Experiments on NPU platforms show approximately $2\times$ throughput
improvement from asynchronous execution, with additional gains from
system-level optimisations, substantially outperforming mainstream RL
frameworks in end-to-end throughput, with speedups of up to
$3\times$ on GPU platforms, further confirming cross-architecture
generalisability while maintaining comparable accuracy.
The proposed framework thus offers a practical, algorithm-agnostic solution
for scalable RL post-training without sacrificing on-policy correctness.
\end{abstract}

\section{Introduction}
Reinforcement learning (RL) has re-emerged as a key technique for
post-training and aligning large language models (LLMs). Following the
introduction of GRPO by DeepSeek-R1~\cite{guo2025deepseek}, RL has
demonstrated strong potential in improving reasoning capabilities, sparking
growing interest in efficient RL pipelines across academia and industry.

Despite these advances, RL for LLMs still faces severe efficiency challenges.
Each training step requires multiple models---the policy, old-policy, and
reference model, and in some cases a value and reward
model~\cite{schulman2017proximal}---and typically depends on generating large
numbers of chain-of-thought (CoT)~\cite{wei2022chain} trajectories using the
latest policy weights, incurring substantial computational and memory overhead.
Early systems such as DeepSpeed-Chat~\cite{yao2023dschat} handled inference
within the policy model directly; subsequent work decoupled training and
inference via dedicated engines~\cite{kwon2023efficient,openr1,hu2024openrlhf}
or adopted Megatron-style 3D parallelism~\cite{feng2025mindspeedrldistributeddataflow,
	sheng2024hybridflow,megatron-lm} to improve throughput. Despite these efforts,
a fundamental bottleneck remains: inference and training execute sequentially
within each step, leaving significant computational resources idle.

Motivated by this, we propose a \emph{periodically asynchronous} framework
that transforms synchronous RL into a producer--consumer pipeline, achieving
concurrent inference and training while remaining strictly on-policy and
requiring no modification to the underlying RL algorithm. We further introduce
a unified tri-model architecture for simultaneous computation of policy,
old-policy, and reference logits, and a shared-prompt attention mechanism to
eliminate redundant computation in long-prompt, short-response settings.

Experiments on NPU platforms show ${\approx}2\times$ throughput improvement
from asynchronous execution alone, with end-to-end speedups of up to
$5.05\times$ over MindSpeed-RL and $3.09\times$ over VERL across NPU and
GPU platforms, while maintaining comparable training effectiveness.
In summary, our contributions are:
(\textbf{i})~a periodically asynchronous RL framework achieving strictly
on-policy asynchronous execution without algorithmic modifications;
(\textbf{ii})~a theoretical proof that the framework remains strictly
on-policy regardless of execution order;
(\textbf{iii})~a unified tri-model architecture with shared-prompt attention
that reduces redundant computation; and
(\textbf{iv})~empirical validation on NPU and GPU platforms demonstrating
substantial throughput improvements with comparable training effectiveness.
\section{Related Work}

We review representative efforts on improving RL efficiency for LLMs, with
emphasis on asynchronous approaches, and situate our work within this
landscape.

\paragraph{The efficiency--correctness trade-off in asynchronous RL.}
A central tension in asynchronous RL system design is the trade-off between
training throughput and on-policy correctness. At the algorithm level,
Asynchronous RLHF~\cite{noukhovitch2024asynchronous} decouples generation and
learning, effectively adopting an online but off-policy strategy that trains on
stale samples from earlier iterations. AReaL~\cite{fu2025areallargescaleasynchronousreinforcement}
advances this by introducing explicit staleness control via a parameter $\eta$,
but doing so requires modifying the training objective itself---replacing
standard PPO with a staleness-aware variant---thereby coupling system-level
efficiency decisions to algorithm-level design. At the system level, ROLL
Flash~\cite{lu2025iirollflash}, Laminar~\cite{sheng2025laminarscalableasynchronousrl},
and LlamaRL~\cite{wu2025llamarldistributedasynchronousreinforcement} propose
fully decoupled architectures that eliminate global synchronisation barriers
and maximise throughput under variable rollout latency. Trajectory Balance with
Asynchrony~\cite{bartoldson2025trajectorybalanceasynchronydecoupling} further
explores decoupling exploration from learning to accelerate post-training.
While these systems achieve substantial throughput gains, they uniformly trade
on-policy correctness for efficiency---either by training on stale samples,
relaxing synchronisation constraints, or introducing algorithm-specific
corrections to compensate for the resulting bias. Critically, their correctness
properties are established only for specific algorithm variants under specific
staleness regimes, leaving their theoretical generalisability across algorithms
and data distributions unclear.

\paragraph{A different design philosophy: asynchrony without algorithmic compromise.}
The on-policy assumption is not merely a theoretical convenience---it
underpins the convergence guarantees, reproducibility, and composability of
standard RL methods. Relaxing it, even with controlled bias, introduces
systemic brittleness whose consequences remain algorithm- and regime-dependent.
Our key insight is that the training iteration itself provides a natural and
sufficient synchronisation boundary: by confining asynchrony to
within-iteration execution and synchronising weights only at iteration
boundaries, we pay no algorithmic price for the efficiency gains we seek.
We therefore ask whether throughput gains can be achieved without sacrificing
on-policy correctness or modifying the underlying RL algorithm, and answer
affirmatively through \emph{periodic asynchrony}: rather than decoupling
inference and training across iterations---which inevitably introduces policy
staleness---we decouple them \emph{within} each iteration while enforcing
strict weight synchronisation at iteration boundaries. This ensures all
rollouts in a batch are generated from the same policy, preserving the
on-policy invariant by construction rather than by approximation. The
accumulated gradient is therefore mathematically identical to that of the
synchronous baseline (Proposition~1 and Remark~1), and no algorithm-specific
correction is required. This design yields three advantages over prior work:
(\textbf{i})~\emph{algorithm-agnosticism}---compatible with any standard
on-policy algorithm, including GRPO and PPO, without staleness-aware variants;
(\textbf{ii})~\emph{formal correctness}---we prove gradient equivalence
regardless of rollout consumption order, a guarantee prior asynchronous
frameworks do not establish; and (\textbf{iii})~\emph{orthogonality}---future
algorithmic improvements can be adopted without re-validating system-level
correctness properties. We further contribute a unified tri-model architecture
and shared-prompt attention that amplify efficiency without compromising these
guarantees, positioning our framework as a correctness-preserving foundation
for scalable RL post-training rather than an efficiency-first approximation
that recovers correctness as an afterthought.
\section{Background}
\label{sec:background}
\subsection*{Training RL with Micro-Batching}
In distributed large-scale training, micro-batching partitions a full batch 
into smaller micro-batches and accumulates gradients before each parameter 
update. This micro-batching strategy is applicable to all PPO-based training algorithms, including GRPO. For a batch of $N$ prompts with $G$ responses each, partitioning 
the $NG$ samples into $M = \lfloor NG/m \rfloor$ micro-batches of size $m$ gives:
\begin{equation}
	J_{\mathrm{batch}}
	= \frac{1}{NG} \sum_{k=1}^{NG}\!\Big(L_k - \beta D_{KL}^k\Big)
	= \frac{1}{M} \sum_{i=1}^{M} \frac{1}{m} \sum_{j=1}^{m}
	\Big(L_{i,j} - \beta D_{KL}^{i,j}\Big),
\end{equation}
where $L_k$ is the PPO-style clipped advantage term and $D_{KL}^k$ is the 
KL penalty between the policy and reference model. The two sides are 
mathematically equivalent, with memory consumption bounded by $m$. This 
property is directly exploited in our asynchronous framework, where samples 
from the inference queue are accumulated as micro-batches and trained 
sequentially without waiting for the full batch.
\section{System Design}

\subsection{System Overview}
\label{sec:overview}
When performing reinforcement learning with large-scale models using
different frameworks for inference and training---for instance, vLLM
for inference and Megatron for training---the process generally involves
three steps: (1)~synchronising the policy model weights with the inference
engine; (2)~the inference engine retrieves prompts from the dataloader,
generates responses, and scores them via a reward module; and (3)~the
generated samples are sent to the training engine for loss computation
and parameter update. Our system preserves this three-step structure
while transforming steps~(2) and~(3) into a concurrent producer--consumer
pipeline.

At a high level, our system decouples inference and training into
independent processes communicating through a shared queue, allowing
both stages to proceed concurrently. The inference process acts as a
producer continuously enqueuing completed rollouts, while the training
process acts as a consumer retrieving rollouts for optimisation as soon
as they become available. Weight synchronisation occurs only at iteration
boundaries, ensuring all rollouts within a batch are generated from the
same policy---the theoretical basis for strict on-policy correctness
established in Section~\ref{sec:correctness}. The following sections
detail the key components of this design.

\subsection{Periodic Asynchronous Reinforcement Learning}
\label{sec:async}
Asynchronous execution of inference and training is a key factor in
accelerating reinforcement learning systems with a separated
training--inference architecture, where the critical aspect is minimising
the idle time the training engine spends waiting for the first completed
rollout to become available. We propose an approach that introduces a
temporary data generator between the data loader and the trainer,
transforming synchronous execution into an asynchronous producer--consumer
pipeline while preserving strict on-policy correctness without any
modifications to the underlying RL algorithm.

\subsubsection{Asynchronous Execution Mechanism}
\label{sec:async_mechanism}

\begin{figure}[ht]
	\centering
	\begin{minipage}[t]{0.49\linewidth}
		\centering
		\includegraphics[width=1.05\linewidth]{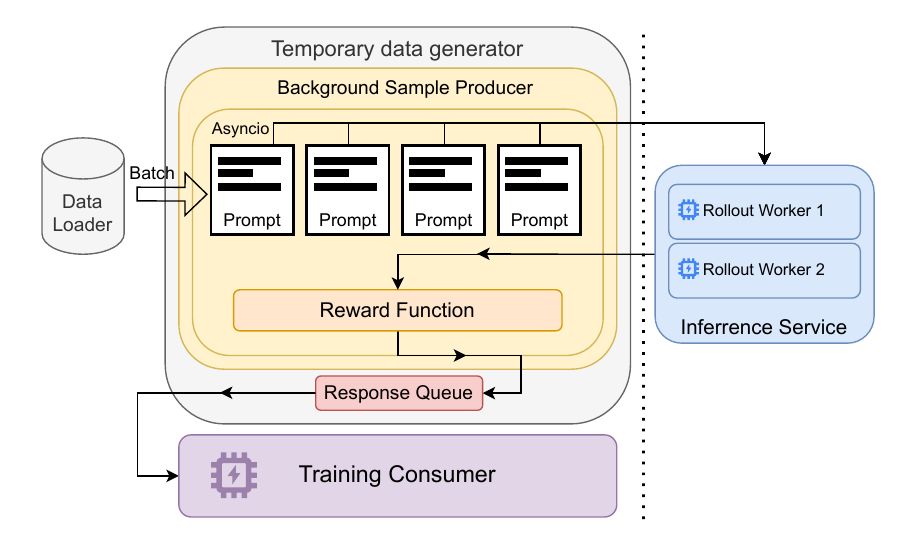}
		\caption{\small Producer–consumer pipeline of the proposed framework: coroutines dispatch prompts to the inference service while the training engine concurrently consumes completed rollouts.}
		\label{fig:async}
	\end{minipage}
	\hfill
	\begin{minipage}[t]{0.49\linewidth}
		\centering
		\includegraphics[width=0.95\linewidth]{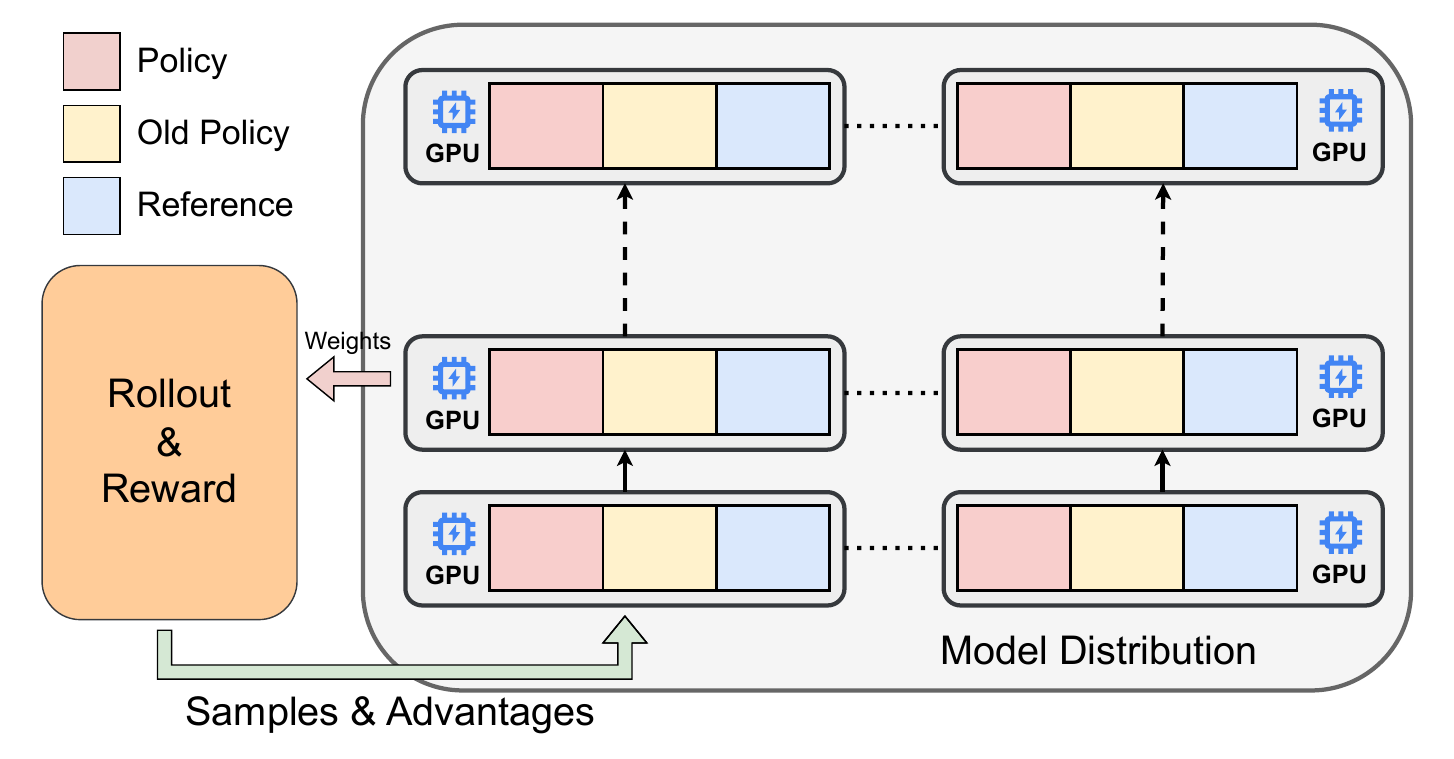}
		\caption{\small Unified tri-model architecture with shared parallel layout, enabling simultaneous computation of policy, old-policy, and reference logits in a single forward pass.}
		\label{fig:arch}
	\end{minipage}
\end{figure}

The reinforcement learning workflow incorporating the temporary sample 
generator is illustrated in Figure~\ref{fig:async}, following a typical 
\emph{producer--consumer} pattern. The pipeline begins with a standard 
\textbf{data source} that loads and provides training prompts in 
batches. Each batch is passed to the \textbf{temporary data 
	generator}—the core component introduced in this work—which runs a 
background thread with parallel coroutines to dispatch prompts to the 
\textbf{inference service} and places the returned rollouts into a 
shared queue. The inference service evenly distributes incoming prompts 
across available instances and processes them efficiently via continuous 
batching. Upon receiving its rollout, each coroutine independently 
evaluates the reward and places it into the queue as well, decoupling 
rollout generation from training optimization.

The main process, acting as the \emph{consumer}, retrieves completed 
rollouts from the queue and feeds them to the \textbf{training engine}. 
Training begins as soon as the first rollout becomes available, allowing 
the training engine to proceed without waiting for the entire batch to 
complete. Once all rollouts in the batch have been consumed, the policy 
is updated and the new weights are synchronized to the rollout workers 
before the next iteration begins. We refer to this design as 
\textbf{periodic asynchrony}, where computation is asynchronous within 
each iteration but synchronized at iteration boundaries.


Each training step requires computing three types of logits—policy, 
old policy, and reference—for every consumed rollout. To enable this, 
we adopt a unified tri-model architecture in which all three models 
share an identical Megatron-style parallel layout with tensor and 
pipeline parallelism. The reference model retains the original weights, 
while the old policy model maintains a one-step delayed copy of the 
policy weights. This shared topology allows all logits to be computed 
simultaneously within a single micro-step (Figure~\ref{fig:arch}). 
After completing the forward passes for a batch, the current policy 
weights are copied to the old policy model \emph{before} the policy 
update is applied (Lines~10--11 in Algorithm~\ref{alg:periodic_async_rl}), 
ensuring that the old policy always retains the weights from the previous 
iteration rather than the current one. This ordering is critical for 
the correctness of the GRPO loss computation, which requires the old 
policy to reflect the distribution under which the rollouts were 
generated. Moreover, the unified design eliminates the need for 
separate resource allocation and scheduling across models, thereby 
simplifying the overall system. The complete procedure is summarized 
in Algorithm~\ref{alg:periodic_async_rl}.

\begin{algorithm}[ht]
	\caption{\small Periodic Asynchronous RL}
	\label{alg:periodic_async_rl}
	\small
	\textbf{Input}: dataset $D$, iterations $T$, batch size $B$ \quad
	\textbf{Output}: trained policy
	\begin{algorithmic}[1]
		\small
		\STATE Initialize shared queue $Q$ for asynchronous communication
		\FOR{$t = 1$ to $T$}
		\STATE Wait until $Q$ is empty, then sync current policy weights $\theta_t$ to rollout workers
		\STATE Sample a batch $P=\{p_i\}_{i=1}^B$ from $D$
		\STATE \textbf{[Background thread]} \textbf{Producer}: for each $p_i$, $r_i\!\gets\!\text{Infer}(p_i)$, $a_i\!\gets\!\text{Reward}(r_i)$; enqueue $(a_i,r_i)$ into $Q$ \hfill $\triangleright$ \textit{runs concurrently with lines 6--9}
		\STATE Initialize accumulated gradient $O=0$
		\FOR{$i = 1$ to $B$}
		\STATE \textbf{[Main thread]} \textbf{Consumer}: dequeue $(a_i,r_i)$ and update $O \gets O + \nabla L(\text{Process}(a_i,p_i,r_i))$
		\ENDFOR
		\STATE Move current policy weights to old policy for stabilization
		\STATE Update policy parameters using accumulated gradient $O$
		\ENDFOR
	\end{algorithmic}
	\hrulefill
\end{algorithm}

\subsubsection{Efficiency Analysis}

\label{sec:async_analysis}

\begin{figure}[!ht]
	\centering
	\hspace{-0.9em}
	\begin{subfigure}[t]{0.49\linewidth}
		\raggedright
		\includegraphics[width=1.15\linewidth]{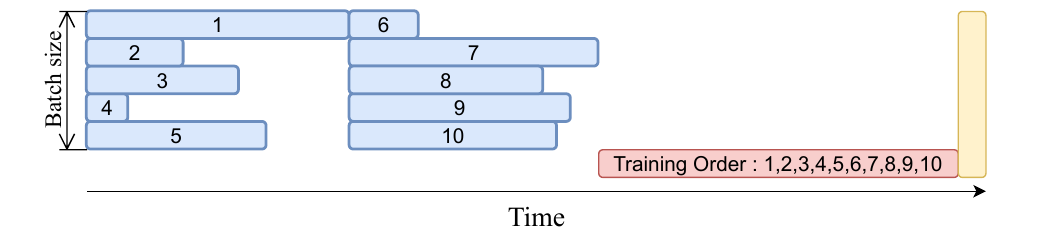}
\caption{\small Synchronous training (single-step iteration)}
		\label{fig:sync_time}
	\end{subfigure}\hspace{-0.08\linewidth}
	\hspace{4.em}
	\begin{subfigure}[t]{0.49\linewidth}
		\centering
		\includegraphics[width=1\linewidth]{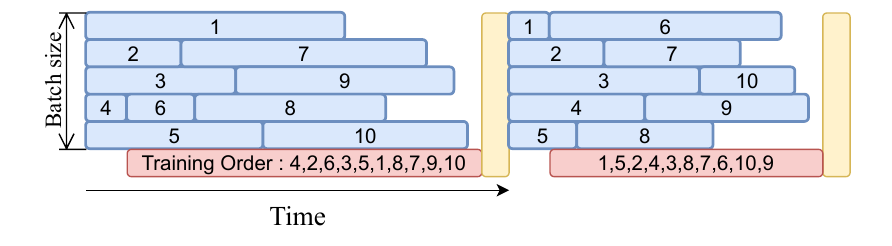}
\caption{\small Asynchronous training (two-step iteration)}
		\label{fig:async_time}
	\end{subfigure}
	
	\vspace{0.5em}
	\includegraphics[width=0.35\linewidth]{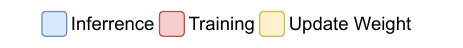}
	
	\caption{\small Wall-clock execution timeline comparing synchronous and asynchronous 
		training. In the synchronous case, training begins only after all rollouts 
		complete; in the asynchronous case, the training engine consumes rollouts as 
		they arrive, enabling inference and training to overlap within each iteration.}
	\label{fig:combined}
\end{figure}

Figure~\ref{fig:sync_time} and Figure~\ref{fig:async_time} illustrate the
wall-clock execution timeline of a single training iteration under each system.
In a synchronous system, inference and training are executed in a strictly
sequential manner: training begins only after all rollouts have completed
inference, and rollouts are consumed in their original prompt order.

The total step time is therefore:
\begin{equation}
	T_{\text{sync}} = T_{\text{infer}} + T_{\text{train}}.
\end{equation}

In the asynchronous system, each completed rollout is immediately enqueued
and consumed by the training worker in completion-time order rather than
the original prompt order. This effectively forms a producer--consumer
pipeline in which inference and training proceed concurrently, reducing
the total step time to:
\begin{equation}
	T_{\text{async}} \approx \max\left\{T_{\text{infer}},\ T_{\text{train}}\right\}.
\end{equation}

The theoretical speedup is thus approximately:
\begin{equation}
	\frac{T_{\text{sync}}}{T_{\text{async}}} 
	\approx \frac{T_{\text{infer}} + T_{\text{train}}}{\max\{T_{\text{infer}},\ T_{\text{train}}\}}
	\leq 2,
\end{equation}
with the upper bound approached when $T_{\text{infer}} \approx T_{\text{train}}$.
This bound corresponds to the ideal efficiency of a two-stage pipeline,
where perfect overlap eliminates idle time between stages.

Furthermore, without continuous batching, synchronous training is gated by
the slowest rollout in each inference batch, introducing additional idle
time. The asynchronous system removes this barrier by enqueuing rollouts
upon completion, and can thus achieve practical speedups exceeding $2\times$.
When inference and training are imbalanced, performance is dominated by the
slower stage; this can be mitigated by reducing training cost (e.g.,
shared-prompt attention, Section~\ref{sec:mask}) and independently scaling
inference and training instances for improved load balancing. Together,
these properties lead to more efficient utilization of compute resources
in heterogeneous environments.

\subsubsection{Correctness Analysis}
\label{sec:correctness}

As shown in Figure~\ref{fig:async_time}, the asynchronous system trains 
samples in completion-time order rather than the original batch order. 
We establish that neither on-policy correctness nor gradient equivalence 
is compromised by this reordering.

\smallskip
\noindent\textbf{Proposition 1} (Periodic Weight Consistency). 
\textit{All rollout samples within the same batch are generated by the 
	same policy $\pi_{\theta_t}$:
	$\forall\, i \in \{1, \ldots, B\}: o_i \sim \pi_{\theta_t}(\cdot \mid p_i)$.}

\noindent\textit{Proof.} Line~3 waits until $Q$ is empty and then 
synchronizes all rollout workers to $\theta_t$ before any rollout in 
the current batch begins. The background producer thread (Line~5) 
dispatches all $B$ prompts exclusively after this synchronization, so 
every $o_i$ is sampled from $\pi_{\theta_t}$. The policy parameters 
are not updated until Lines~10--11, which execute only after all $B$ 
rollouts have been consumed by the main thread (Lines~6--9). Hence 
every $o_i$ is sampled exclusively from $\pi_{\theta_t}$. \hfill$\square$

\smallskip
\noindent\textbf{Remark 1} (Gradient Permutation Invariance).
\textit{The accumulated gradient is invariant to any permutation of the 
	$NG$ training samples, since $\nabla_\theta J_{\mathrm{batch}} = 
	\frac{1}{NG}\sum_{k=1}^{NG} \nabla_\theta(L_k - \beta D^k_{\mathrm{KL}})$ 
	follows directly from the commutativity of finite summation.}

\smallskip
\noindent Together, Proposition~1 and Remark~1 establish that the 
periodic asynchronous framework is strictly on-policy and produces an 
identical parameter update to its synchronous counterpart, with 
convergence guarantees following directly from the underlying RL 
algorithm.

\subsection{Shared-Prompt Attention}
\label{sec:mask}

In GRPO-based reinforcement learning, all samples within a group are 
generated from the same prompt, making it possible to share prompt 
computation across responses within a micro-batch. This optimization 
is most effective when prompts are long relative to responses, where 
redundant prompt recomputation accounts for a significant fraction of 
training cost. The shared-prompt approach introduces four modifications:

\textbf{(1) Input construction.} The shared prompt is concatenated with 
multiple response token IDs as $x = [x_p, x_{r_1}, x_{r_2}]$, with labels 
$y = [y_{r_1}, y_{r_2}]$ excluding the prompt portion.

\textbf{(2) Position indices.} Each response starts immediately after the 
prompt: $p = (0,\dots,|x_p|-1,\ |x_p|,\dots,|x_p|+|x_{r_1}|-1,\ 
|x_p|,\dots,|x_p|+|x_{r_2}|-1)$, where $|\cdot|$ denotes sequence length.

\begin{figure}[ht]
	\centering
	\includegraphics[width=0.4\linewidth]{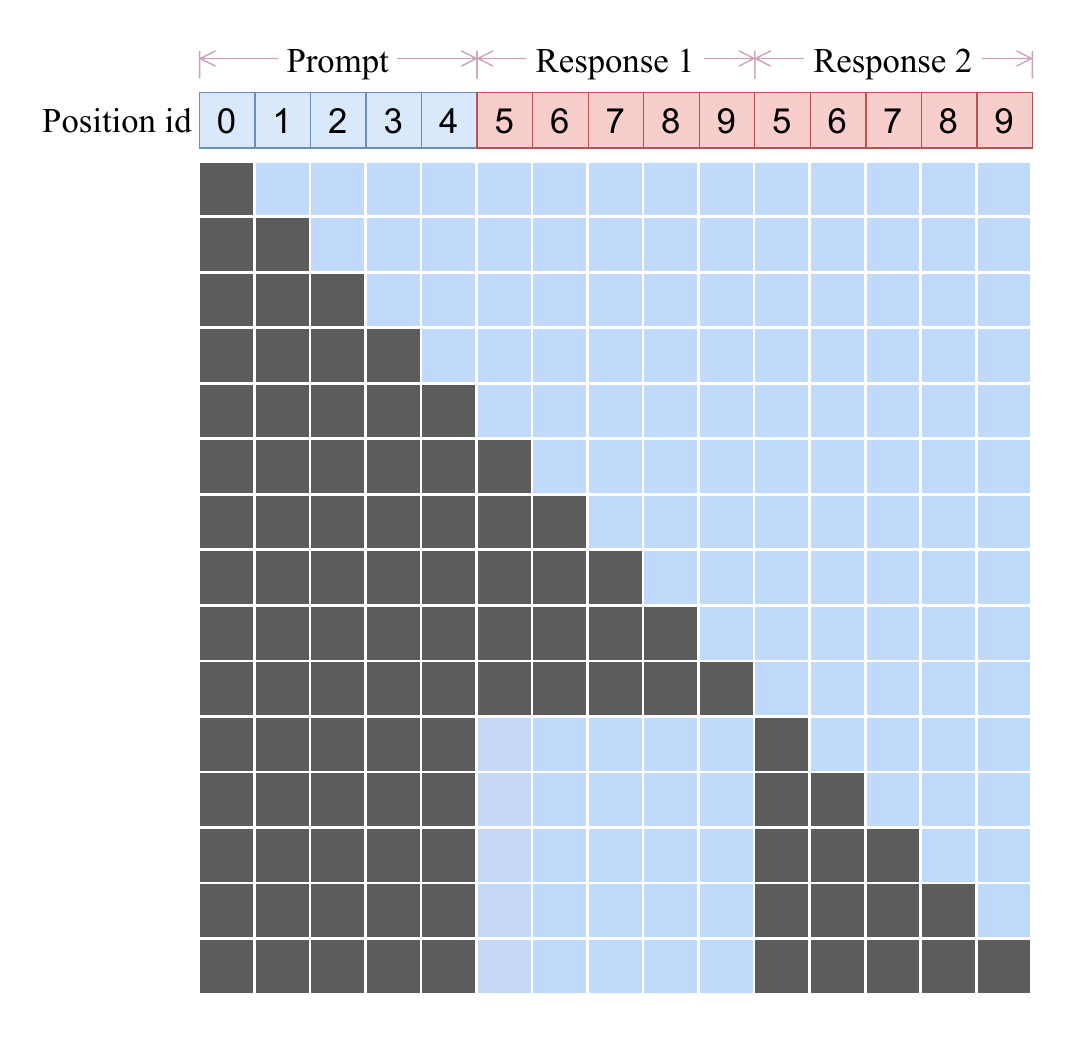}
	\caption{\small Shared-prompt attention mask: each response attends to 
		the shared prompt and its own tokens only, preventing cross-response 
		information leakage while eliminating redundant prompt computation.}
	\label{fig:mask}
\end{figure}

\textbf{(3) Attention mask.} A shared-prompt mask (Figure~\ref{fig:mask}) 
replaces the standard causal mask, restricting each response token to attend 
only to the shared prompt and its own tokens, preventing cross-response 
information leakage.

\textbf{(4) Loss computation.} Prompt tokens are discarded and loss is 
computed only over response tokens: 
$\pi = -\mathrm{CrossEntropy}(\hat{y}_{|x_p|:|x|}, y)$,
where $\hat{y}_{|x_p|:|x|}$ represents the predicted logits of the response 
tokens.

By construction, each response token attends only to the prompt and its 
own preceding tokens, giving $\nabla_\theta \mathcal{L}_{\mathrm{shared}} 
= \sum_{k=1}^{K} \nabla_\theta \mathcal{L}_k$, confirming equivalence to 
standard per-sample training with no approximation or bias. Let $L_p$, 
$L_r$, $K$ denote prompt length, average response length, and responses 
per micro-batch. Standard training redundantly computes prompt tokens $K$ 
times, giving complexity $\mathcal{O}(K(L_p+L_r)^2)$, whereas the 
shared-prompt approach decomposes into prompt self-attention 
$\mathcal{O}(L_p^2)$, response-to-prompt attention $\mathcal{O}(KL_rL_p)$, 
and response self-attention $\mathcal{O}(KL_r^2)$, giving a reduction ratio:
\begin{equation}
	\rho = \frac{L_p^2 + KL_r(L_p+L_r)}{K(L_p+L_r)^2}.
\end{equation}
When $L_p \gg L_r$, $\rho \to \frac{1}{K}$, yielding an approximately 
$K$-fold reduction with no padding overhead, most pronounced in the 
long-prompt, short-response settings common in GRPO reasoning tasks.
\section{Implementation}
\label{sec:Implementation}
Our implementation is built on the \textbf{PyTorch}
training framework, using a 3D-parallel distributed
architecture. We incorporate parts of
\texttt{Megatron-Core} and
\texttt{DeepSpeed}~\cite{rasley2020deepspeed}. On
the \textbf{NPU} platform, we additionally use
\texttt{MindSpeed}, \texttt{torch\_npu}, and
optimized operators such as
\texttt{npu\_fusion\_attention}, an accelerated
attention kernel supporting custom masks and serving
as a counterpart to
\texttt{flash\_attention}~\cite{dao2023flashattention}.
For inference, we use \textbf{vLLM}, with
\texttt{ascend\_vLLM} on NPUs.

As in fine-tuning, all RL samples are computed at
their native sequence lengths for both forward and
backward passes without padding, following
dynamic-length training. The inference and training
components are deployed as separate instances with
a configurable ratio, tuned per platform to balance
throughput.

\section{Experiments}
\label{sec:expt}
We evaluate five reinforcement learning frameworks on mathematical reasoning
tasks: MindSpeed-RL, an official NPU training framework using a Megatron
backend with shared-accelerator design; VERL, a mainstream FSDP-based
framework; AReaL, a fully asynchronous framework that decouples generation
from training via a modified training objective with explicit staleness
control; our synchronous baseline under a decoupled training--inference
design; and our proposed asynchronous framework. Unlike prior work that
validates exclusively on GPU clusters, we evaluate across both NPU and GPU
architectures to confirm hardware-agnosticism. The primary evaluation metric
is end-to-end training throughput, measured by tokens trained per second per
device (TPSPD). Accuracy metrics are provided as reference to verify that no
degradation is introduced by our algorithmic design.

\subsection{Model and Environment Configuration}
\label{sec:model}
The models compared are Qwen2.5-1.5B-Instruct, Qwen2.5-7B-Instruct~\cite{team2024qwen2},
Qwen3-8B~\cite{yang2025qwen3}, and
DeepSeek-R1-Distill-Qwen-32B~\cite{deepseekai2025deepseekr1incentivizingreasoningcapability}.
Training data comes from DeepScaleR~\cite{deepscaler2025} and
GSM8K~\cite{cobbe2021gsm8k}; all experiments use GRPO. For accuracy
evaluation, we employ a rule-based method where the predicted answer is
considered correct if it can be accurately extracted and matches the
ground-truth answer; otherwise it is deemed incorrect. The accuracy test
set is AIME24~\cite{aime24}. NPU experiments were conducted on
\textbf{air-cooled} Ascend-910B NPUs, with each node equipped with eight
64\,GB NPUs, intra-node bandwidth 196\,GB/s and inter-node bandwidth
100\,Gb/s. GPU experiments used a single node of eight NVIDIA A100-40G
GPUs with intra-node bandwidth 64\,GB/s.
\subsection{Training Throughput Comparison}
\label{sec:cmp}

\subsubsection{Comparison with Existing Frameworks}
\paragraph{8B Model on DeepScaleR.}
Table~\ref{tab:main_result} reports results for Qwen3-8B trained on 
DeepScaleR with a 16K context length. Since prompts in this dataset 
are significantly shorter than responses, \textit{Shared-Prompt 
	Attention} is disabled for all methods. Our asynchronous framework 
achieves a TPSPD of 192.259, outperforming MindSpeed-RL 
($\mathbf{3.12\times}$), VERL ($\mathbf{1.24\times}$), and our 
synchronous baseline ($\mathbf{1.92\times}$)---closely approaching 
the theoretical upper bound of $2\times$ derived in 
Section~\ref{sec:async}. Note that VERL's FSDP backend requires a minimum micro-BS of 16, which favors VERL's throughput, making our advantage likely underestimated.

\paragraph{32B Model on DeepScaleR.}
Table~\ref{tab:t2} reports results for DeepSeek-R1-Distill-Qwen-32B. 
In the first group, our asynchronous framework achieves a TPSPD of 
33.449 on only 48 NPUs, delivering a $\mathbf{5.05\times}$ speedup 
over MindSpeed-RL running on 64 NPUs---demonstrating both higher 
throughput and better resource economy. In the second group, 
frameworks are evaluated at 8K context to align with VERL, which 
encounters out-of-memory issues at 16K. The accuracy of our framework 
decreases moderately from 0.738 (16K) to 0.675 (8K), a degradation 
we attribute to response truncation under the reduced context length 
rather than any algorithmic issue; Under the same 8K 
setting, our asynchronous variant achieves a $\mathbf{1.61\times}$ 
speedup over VERL while maintaining comparable accuracy (0.675 vs. 
0.667).

\paragraph{7B Model on GSM8K.}
Table~\ref{tab:shared_prompt} studies the training-dominated regime 
using Qwen2.5-7B-Instruct on GSM8K with a 1K context. Our 
asynchronous framework with \textit{Shared-Prompt Attention} enabled 
achieves a TPSPD of 437.530, corresponding to a $\mathbf{2.20\times}$ 
speedup over MindSpeed-RL and $\mathbf{2.62\times}$ over VERL, while 
both competing frameworks maintain competitive accuracy, confirming 
that the throughput gap is not due to shortcuts in training quality.

\begin{table*}[!htbp]
	\centering
	\begin{minipage}[t]{0.49\textwidth}
		\centering
		\resizebox{0.98\textwidth}{!}{%
			\begin{tabular}{lrrrr}
				\toprule
				Setting & AIME24 & Micro-BS & Training Tokens & TPSPD \\
				\midrule
				Base model (8B)  & 0.725 & --          & --         & --      \\
				MindSpeed-RL     & 0.733 & 1           & 706.355\,M & 61.641  \\
				VERL             & 0.746 & $16\times1$ & 812.084\,M & 155.521 \\
				Sync (ours)      & 0.746 & 1           & 827.306\,M & 99.966  \\
				Async (ours)     & 0.758 & 1           & 917.609\,M & 192.259 \\
				\bottomrule
		\end{tabular}}
		\caption{ 8B model on DeepScaleR, 100 steps, 16 NPUs,
			batch size 32, 32 rollouts per group. $16\times1$: per-GPU
			micro-batch of 1 across 16 NPUs.}
		\label{tab:main_result}
	\end{minipage}
	\hfill
	\begin{minipage}[t]{0.49\textwidth}
		\centering
		\resizebox{0.98\textwidth}{!}{%
			\begin{tabular}{lrrrrr}
				\toprule
				Setting & AIME24 & NPUs & MBS & Training Tokens & TPSPD \\
				\midrule
				Base model (32B) & 0.696 & -- & --          & --         & --     \\
				MindSpeed-RL     & 0.717 & 64 & 1           & 107.874\,M & 6.627  \\
				Sync (ours)      & 0.725 & 48 & 1           & 123.613\,M & 26.219 \\
				Async (ours)     & 0.738 & 48 & 1           & 123.999\,M & 33.449 \\
				\midrule
				VERL             & 0.667 & 64 & $64\times1$ & 164.660\,M & 47.963 \\
				Sync (ours)      & 0.700 & 64 & 1           & 185.796\,M & 46.519 \\
				Async (ours)     & 0.675 & 64 & 1           & 188.081\,M & 77.342 \\
				\bottomrule
		\end{tabular}}
		\caption{32B model on DeepScaleR, 20 steps, 32 rollouts
			per group. Group 1: GBS=32, 16K context.
			Group 2: GBS=64; 8K context due to 16K OOM in VERL.}
		\label{tab:t2}
	\end{minipage}
\end{table*}

\begin{table*}[!htbp]
	\centering
	\begin{minipage}[t]{0.49\textwidth}
		\centering
		\resizebox{0.98\textwidth}{!}{%
			\begin{tabular}{lrrrr}
				\toprule
				Setting & GSM8K & Micro-BS & Training Tokens & TPSPD \\
				\midrule
				Base model (7B)        & 0.801 & --          & --        & --      \\
				MindSpeed-RL           & 0.890 & 16          & 72.534\,M & 199.142 \\
				VERL                   & 0.913 & $16\times1$ & 83.887\,M & 167.297 \\
				Async (ours), w/o SPA  & 0.921 & 1           & 82.655\,M & 52.400  \\
				Sync (ours), w/ SPA    & 0.930 & 16          & 60.586\,M & 218.396 \\
				Async (ours), w/ SPA   & 0.923 & 16          & 60.578\,M & 437.530 \\
				\bottomrule
		\end{tabular}}
		\caption{ 7B model on GSM8K, 16 NPUs, step 116,
			1K context, 32 rollouts. SPA = Shared-Prompt Attention;
			micro-batch 1 = SPA off, 16 = 16 rollouts share one prompt.}
		\label{tab:shared_prompt}
	\end{minipage}
	\hfill
	\begin{minipage}[t]{0.49\textwidth}
		\centering
		\resizebox{0.98\textwidth}{!}{%
			\begin{tabular}{lrrrr}
				\toprule
				Setting & GSM8K & Micro-BS & Training Tokens & TPSPD \\
				\midrule
				Base model (1.5B) & 0.354 & --         & --        & --       \\
				VERL              & 0.782 & $8\times4$ & 85.902\,M & 488.919  \\
				AReaL             & 0.681 & 4          & 88.701\,M & 1067.582 \\
				Sync (ours)       & 0.769 & 4          & 62.671\,M & 628.503  \\
				Async (ours)      & 0.776 & 4          & 62.426\,M & 1510.418 \\
				\bottomrule
		\end{tabular}}
		\caption{ 1.5B model on GSM8K, 8 A100 GPUs.
			All frameworks use data parallelism only, minimizing the impact of inter-device bandwidth on throughput comparison.}
		\label{tab:gpu_result}
	\end{minipage}
\end{table*}

\paragraph{GPU Platform Validation.}
To verify generalizability beyond NPU hardware, we conduct a lightweight 
validation on 8 NVIDIA A100 GPUs using Qwen2.5-1.5B-Instruct on GSM8K, 
comparing against VERL and AReaL. As shown in Table~\ref{tab:gpu_result}, 
our asynchronous variant achieves the highest throughput (TPSPD: 1510.418), 
yielding a $\mathbf{3.09\times}$ speedup over VERL and $\mathbf{1.41\times}$ 
over AReaL. Although AReaL achieves higher throughput than VERL, it incurs 
a substantial accuracy drop (0.681 vs.\ 0.782), possibly attributable to 
off-policy relaxation. In contrast, our method maintains competitive accuracy 
(0.776 vs.\ 0.782) while consuming fewer training tokens. These results 
demonstrate that the proposed framework achieves both throughput gains and 
on-policy correctness consistently across hardware architectures, confirming 
its generalizability beyond NPU platforms.

\subsubsection{Architectural Analysis}
To isolate architectural contributions from asynchronous execution, we 
compare \textit{Sync (ours)} against MindSpeed-RL under identical 
synchronous settings. Our framework achieves a $\mathbf{1.62\times}$ 
speedup on the 8B model (Table~\ref{tab:main_result}) and $\mathbf{3.96\times}$ 
on the 32B model (Table~\ref{tab:t2}), attributable to the decoupled 
training--inference design and unified tri-model architecture. Moreover, 
\textit{Sync (ours), w/ SPA} (TPSPD: 218.396) already surpasses VERL 
(TPSPD: 167.297) in the training-dominated GSM8K setting, demonstrating 
that Shared-Prompt Attention alone is sufficient to close the throughput 
gap even without asynchronous overlap.

\subsubsection{Ablation Analysis}
Table~\ref{tab:shared_prompt} enables a clean ablation 
of the two key components proposed in this work: the 
periodic asynchronous framework and 
\textit{Shared-Prompt Attention}. We isolate their 
respective contributions as follows.

\paragraph{Effect of Shared-Prompt Attention.}
Comparing \textit{Async (ours), w/o SPA} (TPSPD: 
52.400) with \textit{Async (ours), w/ SPA} (TPSPD: 
437.530) under otherwise identical settings reveals 
that enabling \textit{Shared-Prompt Attention} alone 
yields an $\mathbf{8\times}$ improvement in throughput. 
This gain stems from two sources: a reduction in 
training tokens due to shared prompt computation 
(from 82.655\,M to 60.578\,M), and a reduction in 
intra-micro-batch padding overhead arising from 
variable response lengths. These results are consistent 
with the $K$-fold complexity reduction predicted by 
the analysis in Section~\ref{sec:mask}, where $K=16$ 
rollouts share a single prompt.

\paragraph{Effect of Periodic Asynchrony.}
Comparing \textit{Sync (ours), w/ SPA} (TPSPD: 
218.396) with \textit{Async (ours), w/ SPA} (TPSPD: 
437.530) isolates the contribution of asynchronous 
execution under identical architecture and data 
conditions. The asynchronous framework delivers a 
$\mathbf{2\times}$ speedup, closely matching the 
theoretical upper bound established in 
Section~\ref{sec:async}, and confirming that the 
overlap between inference and training is effectively 
exploited in practice. As the two components are 
complementary---\textit{Shared-Prompt Attention} 
reducing per-step training cost and periodic asynchrony 
minimizing idle waiting time---their benefits are 
largely multiplicative, yielding a $\mathbf{2.19\times}$ 
speedup over MindSpeed-RL (Table~\ref{tab:shared_prompt}).

\subsubsection{Accuracy Validation}
\begin{figure}[htbp]
	\centering
	\includegraphics[width=0.95\linewidth]
	{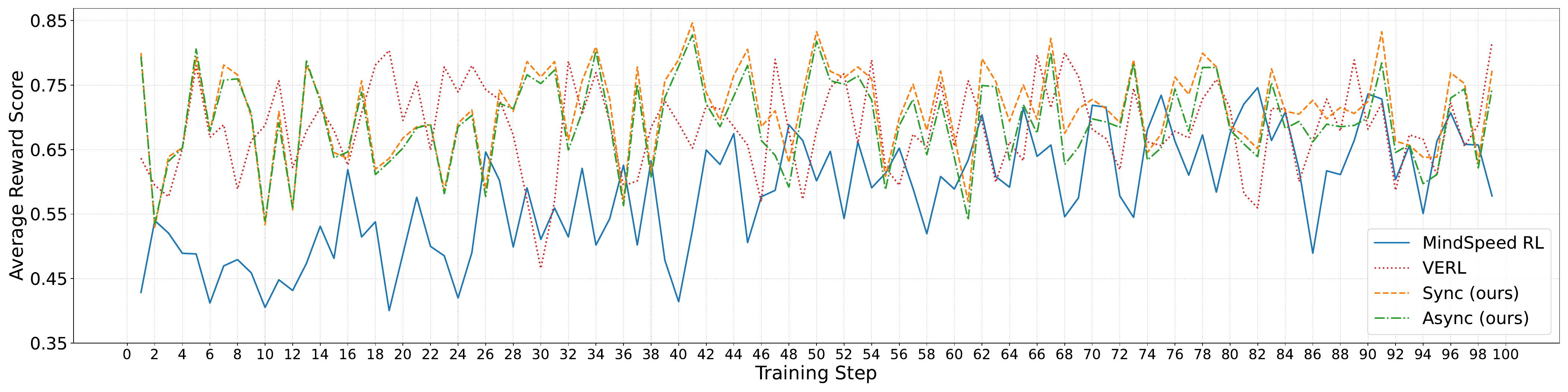}
	\caption{Average reward score across training steps on the 8B model for all frameworks, showing that our synchronous and asynchronous variants maintain comparable training effectiveness throughout.}
	\label{fig:reward_score_comparison}
\end{figure}
As shown across all tables, the accuracy of our 
framework remains consistent with competing methods, 
with absolute differences within 1\% across all 
settings. Furthermore, the reward trajectories of 
our synchronous and asynchronous methods overlap 
throughout training, as shown in 
Figure~\ref{fig:reward_score_comparison}. The 
step-wise reward scores exhibit high variance across 
all frameworks, which is inherent to the discrete 
nature of the rule-based reward function (binary 
correct/incorrect judgment) rather than indicative 
of training instability. Consistent findings are 
also observed in the GPU platform experiments 
(Table~\ref{tab:gpu_result}), where our synchronous 
and asynchronous variants achieve comparable accuracy 
(0.769 vs.\ 0.776), further corroborating this 
conclusion across hardware platforms. Together, these 
results confirm that the substantial throughput gains 
introduced by our framework come at no cost to 
training effectiveness, empirically corroborating 
the on-policy correctness established in 
Proposition~1 and Remark~1.

\subsection{Scalability Analysis}
\label{sec:scale}

\begin{figure}[!htbp]
	\centering
	\raisebox{2em}{
		\begin{minipage}[t]{0.49\linewidth}
			\centering
			\resizebox{0.6\linewidth}{!}{%
				\begin{tabular}{lrrr}
					\toprule
					Model & NPUs & Steps & TPSPD \\
					\midrule
					\multirow{3}{*}{Qwen3-8B}
					& 16 & 20 & 188.162 \\
					& 32 & 20 & 171.824 \\
					& 64 & 20 & 163.208 \\
					\bottomrule
			\end{tabular}}
\captionof{table}{\small Scalability results. TPSPD of 
	Qwen3-8B on 16, 32, and 64 NPUs. 
	Training-to-inference ratio set to 1:4 for optimal throughput.}
			\label{tab:scalability_results}
	\end{minipage}}
	\hfill
	\raisebox{-2em}{
		\begin{minipage}[t]{0.49\linewidth}
			\centering
			\includegraphics[width=0.85\linewidth]{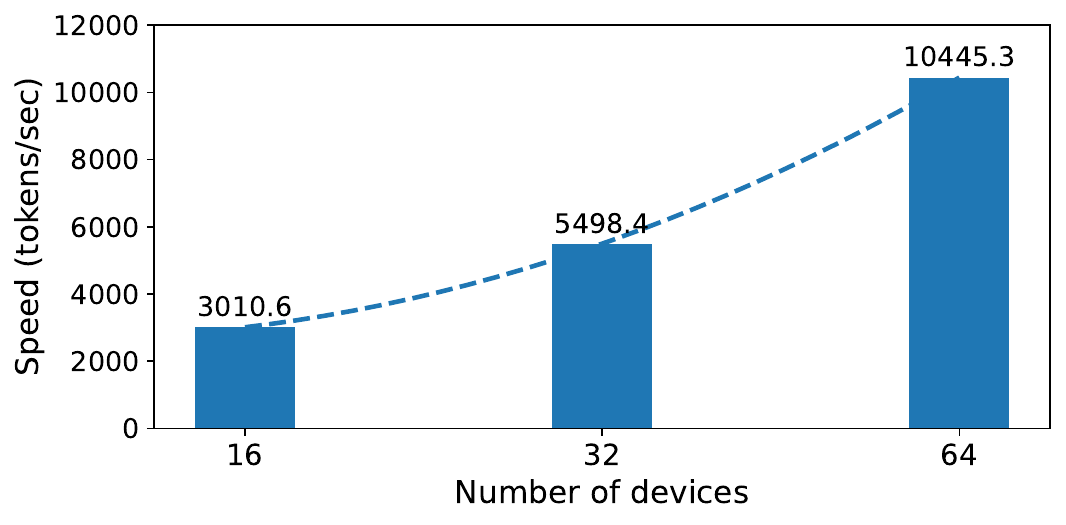}
			\captionof{figure}{\small Total throughput (tokens/sec) at 16, 
				32, and 64 NPUs, demonstrating near-linear scaling.}
			\label{fig:tps_comparison}
	\end{minipage}}
\end{figure}

We conduct a set of experiments to evaluate the 
scalability of our framework using the same 
configuration as the first experiment group. 
The results are shown in 
Table~\ref{tab:scalability_results}.

As shown in Figure~\ref{fig:tps_comparison}, training with 32 NPUs 
achieves a $1.83\times$ speedup over 16 NPUs, and 64 NPUs a $1.9\times$ 
speedup over 32 NPUs, demonstrating near-linear scaling. The moderate 
TPSPD decrease at larger scale is expected due to growing inter-node 
communication overhead.

\section{Conclusion}
\label{sec:conclusion}

This paper addresses the training efficiency bottleneck in on-policy
reinforcement learning by proposing a periodically asynchronous framework
that achieves strictly on-policy asynchronous execution without the
algorithmic modifications required by existing asynchronous approaches.
By introducing a temporary data generator between the data loader and the
trainer, the framework transforms synchronous RL into an asynchronous
producer--consumer pipeline, maximising the overlap between inference and
training. We theoretically establish that the framework remains strictly
on-policy regardless of asynchronous execution order, and introduce a
unified tri-model architecture with a shared-prompt attention mechanism
that significantly reduces redundant computation. Experiments on NPU and
GPU platforms demonstrate around $2\times$ throughput improvement from
asynchronous execution alone, with end-to-end gains of up to $3\times$
over existing frameworks, while maintaining fully comparable training
effectiveness across hardware architectures.

\begin{ack}
\end{ack}

%
\bibliography{neurips_2026}      

@article{guo2025deepseek,
	title={Deepseek-r1 incentivizes reasoning in llms through reinforcement learning},
	author={Guo, Daya and Yang, Dejian and Zhang, Haowei and Song, Junxiao and Wang, Peiyi and Zhu, Qihao and Xu, Runxin and Zhang, Ruoyu and Ma, Shirong and Bi, Xiao and others},
	journal={Nature},
	volume={645},
	number={8081},
	pages={633--638},
	year={2025},
	publisher={Nature Publishing Group UK London}
}

@article{schulman2017proximal,
	title={Proximal policy optimization algorithms},
	author={Schulman, John and Wolski, Filip and Dhariwal, Prafulla and Radford, Alec and Klimov, Oleg},
	journal={arXiv preprint arXiv:1707.06347},
	year={2017}
}

@article{wei2022chain,
	title={Chain-of-thought prompting elicits reasoning in large language models},
	author={Wei, Jason and Wang, Xuezhi and Schuurmans, Dale and Bosma, Maarten and Xia, Fei and Chi, Ed and Le, Quoc V and Zhou, Denny and others},
	journal={Advances in neural information processing systems},
	volume={35},
	pages={24824--24837},
	year={2022}
}

@article{yao2023dschat,
	title={{DeepSpeed-Chat: Easy, Fast and Affordable RLHF Training of ChatGPT-like Models at All Scales}},
	author={Zhewei Yao and Reza Yazdani Aminabadi and Olatunji Ruwase and Samyam Rajbhandari and Xiaoxia Wu and Ammar Ahmad Awan and Jeff Rasley and Minjia Zhang and Conglong Li and Connor Holmes and Zhongzhu Zhou and Michael Wyatt and Molly Smith and Lev Kurilenko and Heyang Qin and Masahiro Tanaka and Shuai Che and Shuaiwen Leon Song and Yuxiong He},
	journal={arXiv preprint arXiv:2308.01320},
	year={2023}
}

@inproceedings{kwon2023efficient,
	title={Efficient Memory Management for Large Language Model Serving with PagedAttention},
	author={Woosuk Kwon and Zhuohan Li and Siyuan Zhuang and Ying Sheng and Lianmin Zheng and Cody Hao Yu and Joseph E. Gonzalez and Hao Zhang and Ion Stoica},
	booktitle={Proceedings of the ACM SIGOPS 29th Symposium on Operating Systems Principles},
	year={2023}
}

@misc{openr1,
	title = {Open R1: A fully open reproduction of DeepSeek-R1},
	url = {https://github.com/huggingface/open-r1},
	author = {{Hugging Face}},
	month = {January},
	year = {2025}
}

@article{hu2024openrlhf,
	title={OpenRLHF: An Easy-to-use, Scalable and High-performance RLHF Framework},
	author={Jian Hu and Xibin Wu and Zilin Zhu and Xianyu and Weixun Wang and Dehao Zhang and Yu Cao},
	journal={arXiv preprint arXiv:2405.11143},
	year={2024}
}

@misc{feng2025mindspeedrldistributeddataflow,
	title={MindSpeed RL: Distributed Dataflow for Scalable and Efficient RL Training on Ascend NPU Cluster}, 
	author={Laingjun Feng and Chenyi Pan and Xinjie Guo and Fei Mei and Benzhe Ning and Jianxiang Zhang and Xinyang Liu and Beirong Zhou and Zeng Shu and Chang Liu and Guang Yang and Zhenyu Han and Jiangben Wang and Bo Wang},
	year={2025},
	eprint={2507.19017},
	archivePrefix={arXiv},
	primaryClass={cs.LG},
	url={https://arxiv.org/abs/2507.19017}, 
}

@article{sheng2024hybridflow,
	title   = {HybridFlow: A Flexible and Efficient RLHF Framework},
	author  = {Guangming Sheng and Chi Zhang and Zilingfeng Ye and Xibin Wu and Wang Zhang and Ru Zhang and Yanghua Peng and Haibin Lin and Chuan Wu},
	year    = {2024},
	journal = {arXiv preprint arXiv: 2409.19256}
}

@inproceedings{rasley2020deepspeed,
	title={Deepspeed: System optimizations enable training deep learning models with over 100 billion parameters},
	author={Rasley, Jeff and Rajbhandari, Samyam and Ruwase, Olatunji and He, Yuxiong},
	booktitle={Proceedings of the 26th ACM SIGKDD international conference on knowledge discovery \& data mining},
	pages={3505--3506},
	year={2020}
}

@article{megatron-lm,
	title={Megatron-LM: Training Multi-Billion Parameter Language Models Using Model Parallelism},
	author={Shoeybi, Mohammad and Patwary, Mostofa and Puri, Raul and LeGresley, Patrick and Casper, Jared and Catanzaro, Bryan},
	journal={arXiv preprint arXiv:1909.08053},
	year={2019}
}

@article{noukhovitch2024asynchronous,
	title={Asynchronous rlhf: Faster and more efficient off-policy rl for language models},
	author={Noukhovitch, Michael and Huang, Shengyi and Xhonneux, Sophie and Hosseini, Arian and Agarwal, Rishabh and Courville, Aaron},
	journal={arXiv preprint arXiv:2410.18252},
	year={2024}
}

@misc{fu2025areallargescaleasynchronousreinforcement,
	title={AReaL: A Large-Scale Asynchronous Reinforcement Learning System for Language Reasoning}, 
	author={Wei Fu and Jiaxuan Gao and Xujie Shen and Chen Zhu and Zhiyu Mei and Chuyi He and Shusheng Xu and Guo Wei and Jun Mei and Jiashu Wang and Tongkai Yang and Binhang Yuan and Yi Wu},
	year={2025},
	eprint={2505.24298},
	archivePrefix={arXiv},
	primaryClass={cs.LG},
	url={https://arxiv.org/abs/2505.24298}, 
}

@misc{lu2025iirollflash,
	title={Part II: ROLL Flash -- Accelerating RLVR and Agentic Training with Asynchrony}, 
	author={Han Lu and Zichen Liu and Shaopan Xiong and Yancheng He and Wei Gao and Yanan Wu and Weixun Wang and Jiashun Liu and Yang Li and Haizhou Zhao and Ju Huang and Siran Yang and Xiaoyang Li and Yijia Luo and Zihe Liu and Ling Pan and Junchi Yan and Wei Wang and Wenbo Su and Jiamang Wang and Lin Qu and Bo Zheng},
	year={2025},
	eprint={2510.11345},
	archivePrefix={arXiv},
	primaryClass={cs.LG},
	url={https://arxiv.org/abs/2510.11345}, 
}

@article{dao2023flashattention,
	title={Flashattention-2: Faster attention with better parallelism and work partitioning},
	author={Dao, Tri},
	journal={arXiv preprint arXiv:2307.08691},
	year={2023}
}

@misc{deepscaler2025,
	title={DeepScaleR: Surpassing O1-Preview with a 1.5B Model by Scaling RL},
	author={Michael Luo and others},
	year={2025},
	howpublished={\url{https://tinyurl.com/deepscaler-2025}},
	note={Notion Blog}
}

@article{yang2025qwen3,
	title={Qwen3 technical report},
	author={Yang, An and Li, Anfeng and Yang, Baosong and Zhang, Beichen and Hui, Binyuan and Zheng, Bo and Yu, Bowen and Gao, Chang and Huang, Chengen and Lv, Chenxu and others},
	journal={arXiv preprint arXiv:2505.09388},
	year={2025}
}

@misc{deepseekai2025deepseekr1incentivizingreasoningcapability,
	title={DeepSeek-R1: Incentivizing Reasoning Capability in LLMs via Reinforcement Learning}, 
	author={DeepSeek-AI},
	year={2025},
	eprint={2501.12948},
	archivePrefix={arXiv},
	primaryClass={cs.CL},
	url={https://arxiv.org/abs/2501.12948}, 
}

@misc{aime24,
	title={American Invitational Mathematics Examination (AIME) 2024}, 
	author={Zhang, Yifan and Math-AI, Team},
	year={2024},
}

@misc{sheng2025laminarscalableasynchronousrl,
	title={Laminar: A Scalable Asynchronous RL Post-Training Framework}, 
	author={Guangming Sheng and Yuxuan Tong and Borui Wan and Wang Zhang and Chaobo Jia and Xibin Wu and Yuqi Wu and Xiang Li and Chi Zhang and Yanghua Peng and Haibin Lin and Xin Liu and Chuan Wu},
	year={2025},
	eprint={2510.12633},
	archivePrefix={arXiv},
	primaryClass={cs.LG},
	url={https://arxiv.org/abs/2510.12633}, 
}

@misc{bartoldson2025trajectorybalanceasynchronydecoupling,
	title={Trajectory Balance with Asynchrony: Decoupling Exploration and Learning for Fast, Scalable LLM Post-Training}, 
	author={Brian Bartoldson and Siddarth Venkatraman and James Diffenderfer and Moksh Jain and Tal Ben-Nun and Seanie Lee and Minsu Kim and Johan Obando-Ceron and Yoshua Bengio and Bhavya Kailkhura},
	year={2025},
	eprint={2503.18929},
	archivePrefix={arXiv},
	primaryClass={cs.LG},
	url={https://arxiv.org/abs/2503.18929}, 
}

@misc{wu2025llamarldistributedasynchronousreinforcement,
	title={LlamaRL: A Distributed Asynchronous Reinforcement Learning Framework for Efficient Large-scale LLM Training}, 
	author={Bo Wu and Sid Wang and Yunhao Tang and Jia Ding and Eryk Helenowski and Liang Tan and Tengyu Xu and Tushar Gowda and Zhengxing Chen and Chen Zhu and Xiaocheng Tang and Yundi Qian and Beibei Zhu and Rui Hou},
	year={2025},
	eprint={2505.24034},
	archivePrefix={arXiv},
	primaryClass={cs.LG},
	url={https://arxiv.org/abs/2505.24034}, 
}

@article{cobbe2021gsm8k,
	title={Training Verifiers to Solve Math Word Problems},
	author={Cobbe, Karl and Kosaraju, Vineet and Bavarian, Mohammad and Chen, Mark and Jun, Heewoo and Kaiser, Lukasz and Plappert, Matthias and Tworek, Jerry and Hilton, Jacob and Nakano, Reiichiro and Hesse, Christopher and Schulman, John},
	journal={arXiv preprint arXiv:2110.14168},
	year={2021}
}

@article{team2024qwen2,
	title={Qwen2 technical report},
	author={Team, Qwen and others},
	journal={arXiv preprint arXiv:2407.10671},
	volume={2},
	number={3},
	year={2024}
}

\appendix

\section{Technical Appendices and Supplementary Material}
\label{app:supplement}
	\section*{Overview}
This document provides detailed reproducibility tables summarizing
the experimental environments, optimization settings, GRPO
hyperparameters, and parallelism configurations used in all
experiments reported in the paper.

\section*{Framework Environments}

\begin{table}[!htbp]
	\centering
	\small
	\begin{tabular}{lcccc}
		\toprule
		\textbf{Framework} & \textbf{Platform} & \textbf{Commit ID}
		& \textbf{Commit Date} & \textbf{vLLM Version} \\
		\midrule
		MindSpeed-RL & NPU & 7741967 & 2025-09-17 & 0.9.1 \\
		VERL         & NPU & 5617529 & 2025-12-09 & 0.11.0 \\
		Ours         & NPU & --      & --         & 0.9.2.rc1 \\
		\midrule
		VERL         & GPU & f9c855f & 2026-01-05 & 0.11.0 \\
		AReaL        & GPU & 3b9eb54 & 2025-12-10 & 0.11.0 \\
		Ours         & GPU & --      & --         & 0.11.0 \\
		\bottomrule
	\end{tabular}
	\caption{Framework versions and environments used in all
		experiments. VERL and AReaL versions used in GPU
		experiments are v0.7.0 and v0.5.0 respectively.}
\end{table}

\section*{Optimization and Precision Settings}

\begin{table}[!htbp]
	\centering
	\small
	\begin{tabular}{>{\raggedright}p{6cm} p{6cm}}
		\toprule
		\textbf{Category} & \textbf{Setting} \\
		\midrule
		Optimizer                & Adam \\
		Learning rate            & $1\times10^{-6}$ \\
		Weight decay             & 0.01 \\
		Gradient norm clipping   & 1.0 \\
		Warmup steps             & 0 \\
		Adam $\beta_1 / \beta_2$ & 0.9 / 0.95 \\
		Gradient checkpointing   & Enabled \\
		\midrule
		Parameter dtype          & bf16 \\
		Gradient dtype           & fp32 \\
		Optimizer state dtype    & fp32 \\
		\bottomrule
	\end{tabular}
	\caption{Shared optimizer and numerical precision settings
		across all frameworks and platforms.}
\end{table}

\section*{GRPO Hyperparameters}

\begin{table}[!htbp]
	\centering
	\small
	\begin{tabular}{l c}
		\toprule
		\textbf{Parameter} & \textbf{Value} \\
		\midrule
		KL coefficient $\beta$  & 0.02 \\
		$\epsilon_{\text{low}}$ & 0.2 \\
		$\epsilon_{\text{high}}$& 0.2 \\
		Answers per prompt      & 32 \\
		Sampling temperature    & 1.0 \\
		Top-p                   & 1.0 \\
		Top-k                   & Disabled \\
		\bottomrule
	\end{tabular}
	\caption{GRPO hyperparameters shared across all experiments.}
\end{table}

\section*{Parallelism and Execution Configuration}

\begin{table}[!htbp]
	\centering
	\small
	\begin{tabular}{l l p{8.5cm}}
		\toprule
		\textbf{Experiment} & \textbf{Framework} & \textbf{Parallelism Configuration} \\
		\midrule
		Experiment 1 & MindSpeed-RL / Ours
		& Actor TP 8, Rollout TP 2 \\
		Experiment 1 & VERL
		& Rollout TP 8, Sequence Parallel 8 \\
		\midrule
		Experiment 2 & MindSpeed-RL / Ours
		& Actor PP 8, Actor TP 8, Rollout TP 4 \\
		Experiment 2 & VERL
		& Rollout TP 8, Sequence Parallel 8 \\
		\midrule
		Experiment 3 & MindSpeed-RL / Ours
		& Actor PP 2, Actor TP 4, Rollout TP 2 \\
		Experiment 3 & VERL
		& Rollout TP 4, Sequence Parallel 1 \\
		
		\midrule
		Experiment 4 (GPU) & VERL
		& Actor/Rollout TP 1, PP 1; training--rollout ratio 1:1 \\
		Experiment 4 (GPU) & AReaL
		& Actor/Rollout TP 1, PP 1; training--rollout ratio 1:1;
		staleness threshold $\eta = 1$ \\
		Experiment 4 (GPU) & Sync (ours)
		& Actor/Rollout TP 1, PP 1; training--rollout ratio 1:1 \\
		Experiment 4 (GPU) & Async (ours)
		& Actor/Rollout TP 1, PP 1; training--rollout ratio 3:1 \\
		\bottomrule
	\end{tabular}
	\caption{Parallelism configurations used in each experiment.
		MindSpeed-RL and VERL adopt coupled training--rollout
		execution, whereas our framework employs a decoupled
		execution design, in which the ratio of training to
		rollout instances is typically set to 1:4; in the
		additional experiments of Experiment~2, this ratio is
		increased to 1:8. In Experiment~4 (GPU), all frameworks
		use data parallelism only (TP$=$1, PP$=$1), with
		configurations tuned to maximize training throughput
		for each framework. AReaL's staleness threshold
		$\eta=1$ is an off-policy-specific parameter
		controlling the maximum number of stale samples
		tolerated per training step.}
\end{table}

\section*{Datasets and Evaluation Settings}

\begin{table}[!htbp]
	\centering
	\small
	\begin{tabular}{l l}
		\toprule
		\textbf{Item} & \textbf{Description} \\
		\midrule
		DeepScaleR       & Training dataset for Experiments 1 and 2 \\
		AIME24           & Validation dataset (30 problems) for Experiments 1 and 2 \\
		GSM8K (train)    & 7,473 samples, used for Experiments 3 and 4 training \\
		GSM8K (test)     & 1,319 samples, used for Experiments 3 and 4 evaluation \\
		\midrule
		Inference engine (NPU)     & vLLM v0.9.2.rc2 \\
		Inference engine (GPU)     & vLLM v0.11.0 \\
		Temperature                & 0.6 \\
		Top-p / Top-k              & 0.95 / 20 \\
		Max tokens (AIME24)        & 32,768 \\
		Max tokens (GSM8K)         & 2,048 \\
		\midrule
		Samples per AIME24 problem & 8 (to reduce evaluation variance) \\
		Samples per GSM8K problem  & 1 \\
		\bottomrule
	\end{tabular}
	\caption{Datasets and inference configurations. Experiment~4
		(GPU) uses the same dataset and evaluation configuration
		as Experiment~3, with a context length of 1K.}
\end{table}

The experimental setup in Experiment~4 (NPU scalability) is
identical to that in Experiment~1, except for the data parallel
size, which is adjusted accordingly. Experiment~4 (GPU) shares
the same dataset, evaluation configuration, and optimization
settings as Experiment~3, with all frameworks restricted to
data parallelism only.

\newpage
\section*{NeurIPS Paper Checklist}

\begin{enumerate}

\item {\bf Claims}
    \item[] Question: Do the main claims made in the abstract and introduction accurately reflect the paper's contributions and scope?
    \item[] Answer: \answerYes{}
\item[] Justification: The abstract and introduction clearly state the four main contributions: the periodically asynchronous framework, the on-policy correctness proof, the unified tri-model architecture with shared-prompt attention, and empirical validation on NPU and GPU platforms. All claims are supported by theoretical results (Propositions 1--2) and experimental results (Tables 1--5).


\item {\bf Limitations}
    \item[] Question: Does the paper discuss the limitations of the work performed by the authors?
\item[] Answer: \answerYes{}
\item[] Justification: The paper acknowledges that throughput gains 
are most pronounced when inference and training times are balanced, 
and that TPSPD decreases moderately with increasing device count due 
to growing inter-node communication overhead (Section~6.3). The 
training-to-inference ratio is a tunable parameter that requires 
per-platform tuning to achieve optimal throughput, as noted in 
Section~5. Additionally, the shared-prompt attention optimization 
is most effective in long-prompt, short-response settings, and 
provides limited benefit when prompts are short relative to responses, 
as discussed in Section~4.3.


\item {\bf Theory assumptions and proofs}
    \item[] Question: For each theoretical result, does the paper provide the full set of assumptions and a complete (and correct) proof?
	\item[] Answer: \answerYes{}
	\item[] Justification: Proposition~1 in Section~4.2.3 provides a 
	complete proof of on-policy correctness, with gradient permutation 
	invariance following as a straightforward remark from the 
	commutativity of finite summation. The correctness and complexity 
	analyses of Shared-Prompt Attention are provided in Section~4.3.

    \item {\bf Experimental result reproducibility}
    \item[] Question: Does the paper fully disclose all the information needed to reproduce the main experimental results of the paper to the extent that it affects the main claims and/or conclusions of the paper (regardless of whether the code and data are provided or not)?
    \item[] Answer: \answerYes{}
\item[] Justification: All hyperparameters, parallelism configurations, hardware specifications, dataset details, and framework versions are reported in Section 6.1 and the supplementary material. Code will be made publicly available upon acceptance.


\item {\bf Open access to data and code}
    \item[] Question: Does the paper provide open access to the data and code, with sufficient instructions to faithfully reproduce the main experimental results, as described in supplemental material?
    \item[] Answer: \answerNo{}
\item[] Justification: To preserve anonymity during review, the code repository is not disclosed at submission time. The code will be released upon acceptance. Training data (DeepScaleR, GSM8K) are publicly available datasets. Detailed reproduction information is provided in the supplementary material.


\item {\bf Experimental setting/details}
    \item[] Question: Does the paper specify all the training and test details (e.g., data splits, hyperparameters, how they were chosen, type of optimizer) necessary to understand the results?
    \item[] Answer: \answerYes{}
\item[] Justification: Section 6.1 details the model configurations, hardware environments, datasets, and evaluation metrics. The supplementary material provides complete optimizer settings, GRPO hyperparameters, parallelism configurations, and dataset/evaluation settings for all experiments.

\item {\bf Experiment statistical significance}
    \item[] Question: Does the paper report error bars suitably and correctly defined or other appropriate information about the statistical significance of the experiments?
 \item[] Answer: \answerYes{}
\item[] Justification: The primary metric TPSPD is a deterministic system throughput measurement that does not require error bars. For accuracy evaluation, AIME24 uses 8 samples per problem averaged to reduce evaluation variance; GSM8K uses 1 sample per problem over 1,319 test problems, where the large test set size provides sufficient stability. These evaluation strategies constitute appropriate measures for statistical reliability given the experimental scale.


\item {\bf Experiments compute resources}
    \item[] Question: For each experiment, does the paper provide sufficient information on the computer resources (type of compute workers, memory, time of execution) needed to reproduce the experiments?
    \item[] Answer: \answerYes{}
\item[] Justification: Section 6.1 specifies the hardware used: Ascend-910B NPUs (64 GB each, 8 per node) for NPU experiments and NVIDIA A100-40G GPUs (8 per node) for GPU experiments, along with intra- and inter-node bandwidth details.

    
\item {\bf Code of ethics}
    \item[] Question: Does the research conducted in the paper conform, in every respect, with the NeurIPS Code of Ethics \url{https://neurips.cc/public/EthicsGuidelines}?
    \item[] Answer: \answerYes{}
\item[] Justification: This work proposes a training efficiency framework for LLMs and does not involve human subjects, sensitive data, or applications with direct negative societal impact.

\item {\bf Broader impacts}
    \item[] Question: Does the paper discuss both potential positive societal impacts and negative societal impacts of the work performed?
    \item[] Answer: \answerNA{}
\item[] Justification: This paper focuses on improving the computational efficiency of RL training for LLMs. It is foundational infrastructure research with no direct path to specific negative applications beyond those already associated with LLMs in general.

    
\item {\bf Safeguards}
    \item[] Question: Does the paper describe safeguards that have been put in place for responsible release of data or models that have a high risk for misuse (e.g., pre-trained language models, image generators, or scraped datasets)?
    \item[] Answer: \answerNA{}
\item[] Justification: This paper proposes a training framework and does not release new models or datasets that pose high risk for misuse.

\item {\bf Licenses for existing assets}
    \item[] Question: Are the creators or original owners of assets (e.g., code, data, models), used in the paper, properly credited and are the license and terms of use explicitly mentioned and properly respected?
    \item[] Answer: \answerYes{}
\item[] Justification: All datasets (DeepScaleR, GSM8K, AIME24), models (Qwen2.5, Qwen3, DeepSeek-R1-Distill), and frameworks (vLLM, Megatron-Core, DeepSpeed, MindSpeed) are properly cited. All are publicly released under open licenses.

\item {\bf New assets}
    \item[] Question: Are new assets introduced in the paper well documented and is the documentation provided alongside the assets?
    \item[] Answer: \answerYes{}
\item[] Justification: The proposed framework is documented in the paper and supplementary material. The code will be publicly released upon acceptance. No new datasets or models are introduced.

\item {\bf Crowdsourcing and research with human subjects}
    \item[] Question: For crowdsourcing experiments and research with human subjects, does the paper include the full text of instructions given to participants and screenshots, if applicable, as well as details about compensation (if any)? 
    \item[] Answer: \answerNA{}
\item[] Justification: This paper does not involve crowdsourcing or research with human subjects.

\item {\bf Institutional review board (IRB) approvals or equivalent for research with human subjects}
    \item[] Question: Does the paper describe potential risks incurred by study participants, whether such risks were disclosed to the subjects, and whether Institutional Review Board (IRB) approvals (or an equivalent approval/review based on the requirements of your country or institution) were obtained?
    \item[] Answer: \answerNA{}
\item[] Justification: This paper does not involve human subjects research.

\item {\bf Declaration of LLM usage}
    \item[] Question: Does the paper describe the usage of LLMs if it is an important, original, or non-standard component of the core methods in this research? Note that if the LLM is used only for writing, editing, or formatting purposes and does \emph{not} impact the core methodology, scientific rigor, or originality of the research, declaration is not required.
    \item[] Answer: \answerNA{}
\item[] Justification: LLMs are the subject of study rather than a tool used in the research methodology. No LLMs were used in a non-standard way for writing, experimental design, or data analysis.

\end{enumerate}
\end{document}